\documentclass[10pt,twocolumn,letterpaper]{article}

\usepackage[pagenumbers]{cvpr}

\usepackage{cuted} 

\usepackage{currfile} 

\usepackage{caption} 

\usepackage{amsthm}
\newtheorem{proposition}{Proposition}
\usepackage{algorithm}
\usepackage{algpseudocode}

\usepackage{pifont}
\usepackage{microtype}
\usepackage{balance} 

\makeatletter
\def\@maketitle{%
  \newpage
  \null
  \vspace*{-21pt}%
  \begin{center}
    \hrule height 5pt\relax
    \vspace{16pt}%
    {\fontsize{19}{22}\selectfont\bfseries\@title\par}%
    \vspace{17pt}%
    \hrule height 0.8pt\relax
    \vspace{28pt}%
    {\large\begin{tabular}[t]{c}\@author\end{tabular}\par}%
    \vspace{9pt}%
  \end{center}%
}
\makeatother

\definecolor{cvprblue}{rgb}{0.21,0.49,0.74}
\usepackage[breaklinks,colorlinks,allcolors=cvprblue]{hyperref}
\hypersetup{
  pdftitle={SOAR: Training-Free Rank-Consistent Rollout Compression for Latent World Model Planning},
  pdfauthor={Anqi Li, Kai Liu, Linghe Kong, Yulun Zhang}
}

\def\confName{CVPR}
\def\confYear{2027}

\title{SOAR: Training-Free Rank-Consistent Rollout\\Compression for Latent World Model Planning}

\author{%
{\bfseries Anqi Li\textsuperscript{1,2,*}\quad
Kai Liu\textsuperscript{2,*}\quad
Linghe Kong\textsuperscript{2}\quad
Yulun Zhang\textsuperscript{2}}\\[3pt]
{\normalsize \textsuperscript{1}Zhiyuan College, Shanghai Jiao Tong University\qquad
\textsuperscript{2}Shanghai Jiao Tong University}\\[2pt]
{\small \textsuperscript{*}These authors contributed equally.}
}

\begin{document}
\raggedbottom 
\maketitle
\definecolor{figgrey}{HTML}{7F7F7F}
\definecolor{figblue}{HTML}{1F5FAE}
\definecolor{figorange}{HTML}{E8742A}
\begin{strip}
\centering
\includegraphics[width=\textwidth]{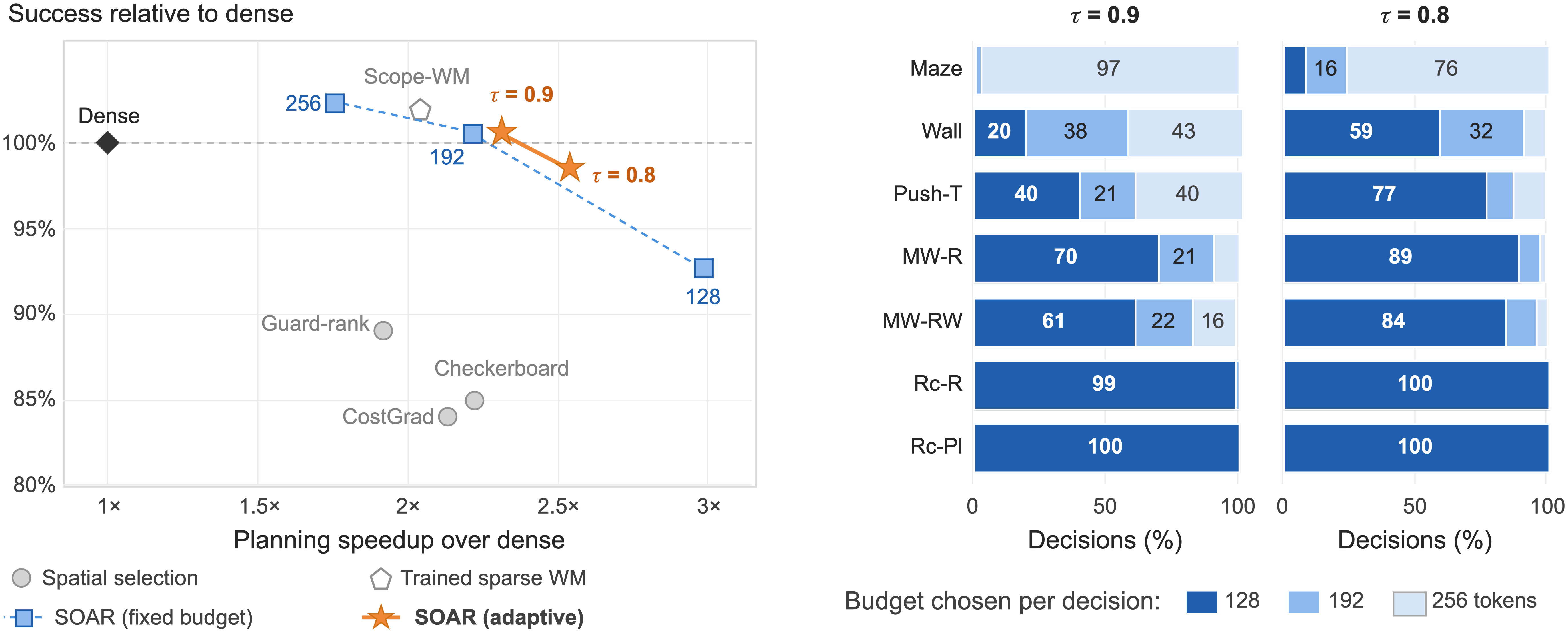}
\captionsetup{hypcap=false}
\captionof{figure}{\textbf{Planning needs the ranking, not the reconstruction.}
\emph{Left:} success and speedup relative to the dense model of the same batch. Choosing which tokens to keep saturates below dense (\textcolor{figgrey}{\textbf{grey}}); an exact first step and a cached context move the frontier (\textcolor{figblue}{\textbf{blue}}); letting the candidate ranking pick each decision's budget reaches dense-level success at $2.3\times$ (\textcolor{figorange}{\textbf{orange}}).
\emph{Right:} share of decisions at which SOAR computes 128, 192 or 256 tokens of the newest frame: most compression on MetaWorld and RoboCasa, least on Maze, and varying across decisions on Push-T and Wall. One-step-horizon decisions count as 256.}
\captionsetup{hypcap=true}
\label{fig:teaser}
\end{strip}

\begin{abstract}
Latent world models plan by rolling out hundreds of candidate action sequences through a transformer predictor, so a single decision can take tens of seconds.
Existing accelerations prune spatial tokens as in vision-language models, yet a sampling-based planner uses predicted costs only to rank candidates, so a rollout can be compressed freely as long as the ranking survives.
We propose SOAR, a training-free method that adapts rollout compression to each planning decision.
At every decision, SOAR probes a few candidates under several token budgets and keeps the smallest budget that preserves their ranking, so it compresses as far as the current decision allows, without task labels.
A closed-loop study of more than 80{,}000 episodes fixes the rest of the rollout: the first step, which alone reads a real observation, stays exact, and the context frame is cached rather than recomputed.
On seven navigation and manipulation tasks with frozen checkpoints, SOAR matches dense-level planning success at $2.3\times$ speedup and retains 98\% at $2.5\times$.
Our code is available at \url{https://github.com/starrywiki/SOAR}.
\end{abstract}

\section{Introduction}
\label{sec:intro}

Latent world models~\cite{zhou2025dinowm,assran2025vjepa2,terver2025jepawms} plan by imagination.
At every decision, the cross-entropy method (CEM)~\cite{rubinstein2004cem} samples hundreds of action sequences, rolls each one out autoregressively through a transformer predictor over a dense grid of patch features, and keeps the candidates with the lowest predicted cost.
With the original budget, one planning call of the released JEPA-WM~\cite{terver2025jepawms} takes 20--50 seconds on an H100 and recurs at each replanning step.

The natural remedy, and the one recent work has adopted, is token reduction.
Sparse Imagination~\cite{chun2026sparse}, DDP-WM~\cite{yin2026ddpwm}, Scope-WM~\cite{li2026scopewm} and CostGrad~\cite{xu2026costgrad} all decide \emph{which spatial tokens} the predictor should process in full.
This framing is inherited from vision-language models~\cite{bolya2023tome,chen2024fastv,wen2025dart,alvar2025divprune}, where an image is encoded once and the kept tokens must preserve its content.
A planner differs in what it consumes.
A vision-language model consumes the tokens themselves, whereas CEM consumes only an ordering: it keeps the $K$ lowest-cost candidates and discards the costs.
A cheaper rollout is therefore exactly as good as the dense one whenever it selects the same elites, however much its predicted features differ.
This changes the question.
Token selection asks which tokens best preserve a frame; a planner asks which computation, if removed, reorders the candidates.
Because the predictor consumes its own outputs, the answer need not be spatial: an error at the first step is inherited by every later prediction and by the cost; one at the last step is not.

We therefore take the candidate ranking as the criterion and ask along which axis a latent world-model rollout can be reduced without reordering the candidates.
We prune along four axes at comparable compute---space, the context frame, the rollout step and predictor depth---using frozen released checkpoints, the original planner budget, and more than 80{,}000 closed-loop episodes on seven navigation and manipulation tasks.
Three findings emerge, and together they determine the design of our method.
\begin{itemize}
\item \textbf{Which tokens matters little.} At a 50\% budget, rules from vision-language models are near or below random selection, and planning-aware selectors stop 5 to 8 points below the dense model, within a few points of a fixed checkerboard, as shown in \cref{sec:spatial}.
\item \textbf{When matters: the first step anchors the ranking.} Step~0 is the only step that reads the real observation. Computing it in full and compressing only later steps recovers 5.4 points lost to spatial pruning, while spending comparable compute on the last step instead drops success to 0.66 of dense, as shown in \cref{sec:axes}.
\item \textbf{Context must be kept but not recomputed.} After step~0, half of the predictor's work recomputes the previous frame only to provide keys and values. Dropping that frame is free on navigation and pushing but costs 12 to 43 points on robot-arm reaching; reusing its cached keys and values removes the trade-off and matches dense success at a $1.76\times$ speedup without knowing the task, as demonstrated in \cref{sec:context,sec:experiments}.
\end{itemize}

The findings also reveal that tasks differ in \emph{what} they need: halving the spatial positions costs Push-T 27\% of its dense successes but leaves MetaWorld Reach unchanged, whereas dropping the context frame does the opposite, as illustrated in \cref{fig:analysis}b.
Once the context is cached, the only remaining question is how far the newest frame can be compressed, which varies across tasks and individual decisions.

We propose \textbf{SOAR}, a training-free method that adapts rollout compression online, choosing at every decision how much computation the candidate ranking needs.
At each decision, SOAR rolls out 16 candidates of the first CEM population under three nested token budgets and selects the smallest budget whose Spearman rank correlation with the full budget reaches a threshold, so each decision is compressed as far as its ranking allows.
This adaptive choice is affordable because the findings fix everything else: SOAR computes the first step in full and reads the context frame from a cache, which leaves a single scalar per decision, the token budget of the newest frame.
The probe adds 1--2\% to the planning time and requires neither task labels nor training.
Without being told the task, SOAR compresses most on MetaWorld and RoboCasa, least on Maze, and differently across decisions on Push-T and Wall, as shown in \cref{fig:teaser}. It reaches dense-level success at a $2.3\times$ speedup, whereas every cheaper fixed budget loses successes on tasks that need spatial detail, as detailed in \cref{sec:experiments}.

\paragraph{Contributions.}
\textbf{Online criterion.} We recast rollout compression for sampling-based planning as preserving the elite ranking rather than the predicted features, as established by \cref{prop:rank}, and show that this criterion is cheap enough to check at every decision: 16 probe candidates, 1--2\% of the planning time.
\textbf{Diagnosis.} In a closed-loop study along four axes at comparable compute on frozen released checkpoints, we find that spatial selection saturates regardless of the selector, that the first step anchors the ranking, that the context frame must be kept but not recomputed, and that tasks fail along different axes, so no fixed budget is safe for all of them.
\textbf{Method.} SOAR keeps what the diagnosis shows must stay exact and lets the ranking decide the rest, reaching dense-level success at a $2.3\times$ speedup and 98\% of it at $2.5\times$ with no training and no task label, while avoiding the task-specific failures of cheaper fixed budgets.

\section{Related Work}
\label{sec:related}

\paragraph{Token reduction for vision transformers and VLMs.}
Token merging~\cite{bolya2023tome} combines similar tokens inside ViTs, FastV~\cite{chen2024fastv} drops visual tokens after the early layers of a vision-language model based on attention scores, DART~\cite{wen2025dart} removes duplicated tokens, and DivPrune~\cite{alvar2025divprune} keeps a diverse subset.
These methods accelerate a single forward pass, whereas a planner calls the predictor repeatedly inside a search and uses only the ranking of the resulting costs; rules that switch token sets inside the network, which work well in VLMs, are harmful in planning, as shown in \cref{sec:axes}.

\paragraph{Efficient world-model planning.}
Latent world models such as DINO-WM~\cite{zhou2025dinowm}, V-JEPA~2-AC~\cite{assran2025vjepa2} and JEPA-WMs~\cite{terver2025jepawms} plan by optimizing actions with the cross-entropy method~\cite{rubinstein2004cem} over predicted patch features.
Several methods reduce the number of spatial tokens the predictor processes.
Sparse Imagination~\cite{chun2026sparse} trains a predictor on random token subsets and draws a new random mask at each planning step; CostGrad~\cite{xu2026costgrad} selects tokens without training from gradients of the planning cost and keeps the subset fixed across candidates and CEM iterations of a decision; DDP-WM~\cite{yin2026ddpwm} and Scope-WM~\cite{li2026scopewm} train predictors that localize the dynamic tokens of each predicted frame and update the remaining ones with a cheaper module.
These methods differ in how tokens are selected, but all reduce computation in space.
We instead ask \emph{along which axis} of the rollout redundancy lies, and show that the answer depends more on when computation is spent than on which tokens are kept.

\paragraph{Adaptive computation in planning.}
A few works adapt how much computation is spent.
Sparse Imagination includes a variant that raises or lowers its drop ratio between CEM iterations according to the variance of latent rollouts~\cite{chun2026sparse}; CF-Prune~\cite{kim2026cfprune} prunes future-frame tokens in diffusion-based world action models and keeps dense computation during phases it detects as precision-critical; SufficientPlan~\cite{li2026sufficientplan} certifies a reduced search budget for each model--task pair from paired closed-loop evaluations and caches static context across search iterations.
SOAR differs in its signal and its granularity: it selects the token budget at each decision from the agreement of candidate rankings, the quantity a sampling-based planner actually consumes, rather than from prediction uncertainty or offline closed-loop evaluation, with a frozen, dense-trained predictor.

\paragraph{Caching in autoregressive video and world models.}
Reusing the keys and values of past frames is standard in autoregressive generation.
M*~\cite{jha2026mstar} applies KV caching to V-JEPA~2-AC planning rollouts, while TempCache~\cite{samuel2026tempcache}, WorldKV~\cite{yi2026worldkv} and X-Cache~\cite{zeng2026xcache} compress, retrieve or skip cached computation in video world models.
Caching is an existing building block; we show that it resolves a pruning trade-off because some tasks need the context frame and others do not, as detailed in \cref{sec:context}.

\paragraph{Temporal context in world models.}
VLWM~\cite{du2026vlwm} notes strong redundancy between adjacent frames and replaces step-by-step rollouts with variable-length jumps, and ActSWM~\cite{gan2026actswm} reports that richer visual context can make predictors insensitive to actions.
Our results refine both for planning: whether the previous frame matters depends on the task dynamics, and when it does, coverage of the frame matters more than its motion.

\section{Preliminaries and Evaluation Protocol}
\label{sec:setup}

\paragraph{World model and planner.}
We use the released JEPA-WM checkpoints~\cite{terver2025jepawms}.
A frozen pretrained DINOv2 ViT-S/14 encoder maps each image to a $16\times16$ grid of patch features; RoboCasa instead uses DINOv3 ViT-L/16.
An action-conditioned AdaLN transformer predicts the next feature grid; it has 6 layers and width 384 for Maze, Wall, Push-T and MetaWorld, and 12 layers and width 1024 for RoboCasa.
Planning uses CEM with $N{=}300$ candidates and $K{=}10$ elites for 30 iterations on Maze, Wall and Push-T or 15 on MetaWorld and RoboCasa. The horizon $H$ is 6 or 3 steps, and the L2 cost compares the final predicted features, including proprioception when available, with the goal.

\paragraph{What a planner needs from a rollout.}
Let $c(a)$ be the cost of candidate $a$ under the dense rollout and $\tilde c(a)$ its cost under a cheaper rollout.
Each CEM iteration only refits its sampling distribution to the elite set $\mathcal{E}=\operatorname{top}_K\{-c(a_i)\}_{i=1}^{N}$.
A reduced rollout therefore leaves the planner unchanged whenever it selects the same elites, even if $\tilde c$ differs greatly from $c$ in value.
This is a weaker requirement than faithful feature prediction, and it is the lens we use throughout: a reduction is harmful when it reorders candidates, not when it changes the predicted features.

\begin{figure}[t]
\centering
\includegraphics[width=\linewidth]{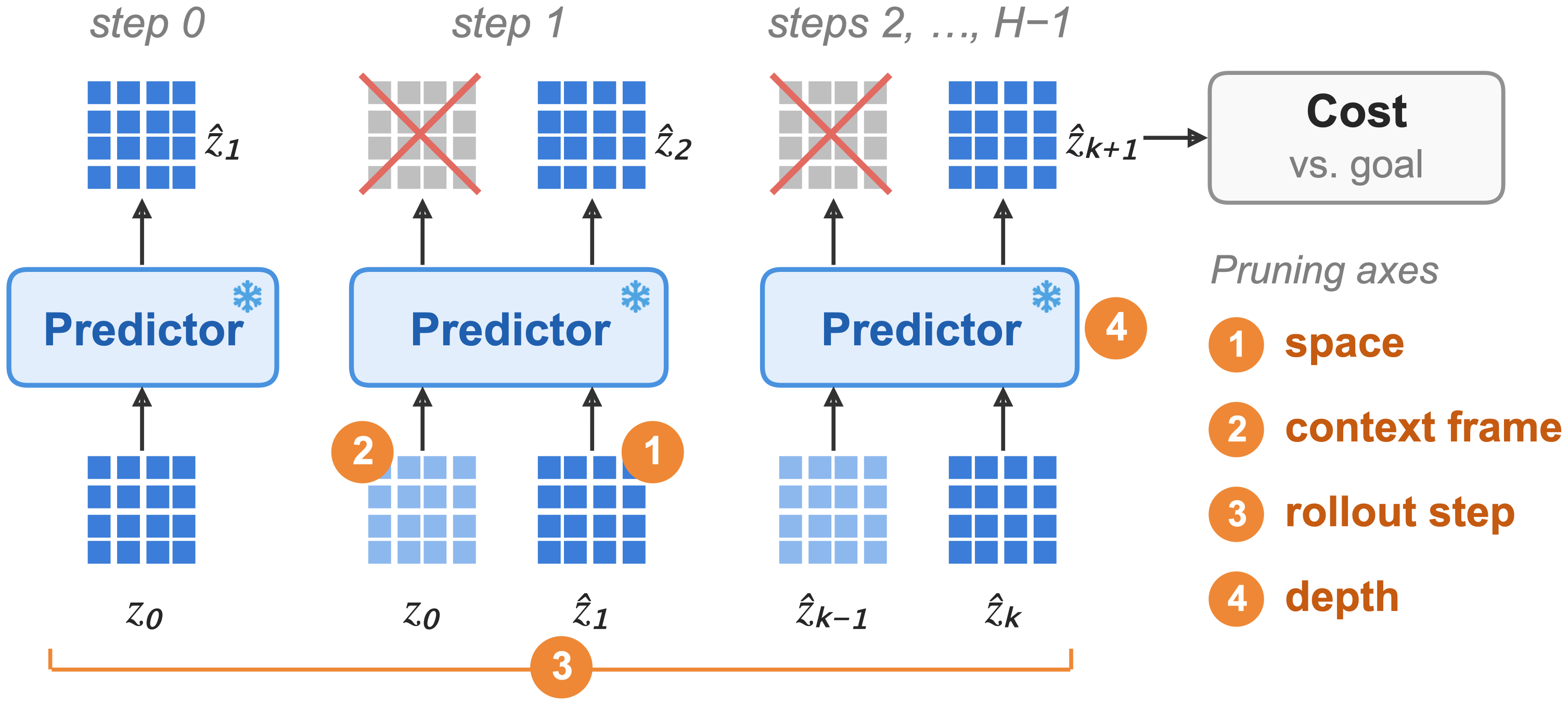}
\caption{\textbf{Computation in a planning rollout.} Step~0 reads the encoded observation. Every later step recomputes a two-frame window, but only the newest frame's output is kept; the context frame is recomputed solely to provide keys and values, and its output is discarded. Numbers mark the pruning axes of \cref{sec:analysis}.}
\label{fig:anatomy}
\end{figure}

\paragraph{Anatomy of a rollout.}
The predictor attends over a sliding window of two frames.
At step~0 it reads the observation; from step~1 on it reads the previous frame and the newest frame, recomputes both, and keeps only the newest frame's output, as illustrated in \cref{fig:anatomy}.
For a horizon $H$, a candidate rollout therefore processes $256 + 512(H-1)$ tokens per layer, and half of the work after step~0 is spent on a context frame whose outputs are discarded.
We count compute as tokens times layers relative to the dense rollout and report the measured speedup alongside it.
Unless stated otherwise, pruned positions are removed from the state, and the cost is evaluated on the retained positions of the final frame, with the goal features restricted to the same positions.
Keeping stale positions barely changes success: 0.89 versus 0.90.

\paragraph{Tasks and closed-loop evaluation.}
We evaluate on Maze, Wall, Push-T, MetaWorld Reach (MW-R) and Reach-Wall (MW-RW), and RoboCasa Reach (Rc-R) and Place (Rc-Pl).
Each evaluation runs 96 episodes on the first five tasks and 32 on RoboCasa, with the original episode lengths and replanning frequency.
All methods share initial states and goals, which gives paired comparisons.
We report the task-averaged success rate relative to the dense model evaluated in the same batch, and 95\% intervals from a paired bootstrap stratified by task using 10{,}000 resamples.
Latency is the wall-clock time of one complete planning call, including encoding, every auxiliary computation and the full CEM search, measured on a single H100 in FP32 and divided by the dense latency on the same task; we report its inverse as the speedup. SOAR reduces the number of tokens processed and is orthogonal to lower-precision arithmetic and kernel compilation, which reduce the cost per token.
Every method is compared with a dense model run on the same episodes in the same batch. \Cref{tab:spatial,tab:axes,tab:main} report success relative to this dense model, and \cref{tab:soar} reports absolute successes.
In total, the study comprises more than 80{,}000 closed-loop episodes.

\section{Where Is a Planning Rollout Redundant?}
\label{sec:analysis}

We study where a rollout of the frozen predictor can be reduced without hurting closed-loop planning: by examining spatial token selection in \cref{sec:spatial}, the timing of reduction in \cref{sec:axes}, and the role of the context frame in \cref{sec:context}.

\subsection{Spatial token selection saturates}
\label{sec:spatial}

We first apply the standard recipe: keep 128 of the 256 spatial positions per frame for the whole rollout and choose them with a selection rule.
We adapt token merging (ToMe)~\cite{bolya2023tome}, attention-based FastV~\cite{chen2024fastv}, duplication-based DART~\cite{wen2025dart} and diversity-based DivPrune~\cite{alvar2025divprune}, and evaluate CostGrad~\cite{xu2026costgrad} and Guard-rank, a probe-guided variant that picks recipes by rank agreement with the dense model.

\begin{table}[t]
\centering
\small
\setlength{\tabcolsep}{6pt}
\begin{tabular}{lccc}
\toprule
Method & Success / Dense & $\Delta$ (points) & Speedup \\
\midrule
Dense & 1.00 & -- & 1.00$\times$ \\
Random & 0.60 & $-$19.6 & 2.13$\times$ \\
\midrule
\multicolumn{4}{l}{\emph{From vision-language models}} \\
ToMe~\cite{bolya2023tome} & 0.51 & $-$24.4 & 2.13$\times$ \\
FastV~\cite{chen2024fastv} & 0.66 & $-$16.6 & 1.35$\times$ \\
DART~\cite{wen2025dart} & 0.34 & $-$32.7 & 1.52$\times$ \\
DivPrune~\cite{alvar2025divprune} & 0.63 & $-$18.1 & 2.00$\times$ \\
\midrule
\multicolumn{4}{l}{\emph{Planning-aware}} \\
CostGrad~\cite{xu2026costgrad} & 0.84 & $-$7.9 & 2.13$\times$ \\
Guard-rank & 0.89 & $-$5.4 & 1.92$\times$ \\
\bottomrule
\end{tabular}
\caption{\textbf{Spatial token selection at a 50\% budget.} Seven-task average success relative to dense, difference in points, and speedup, with the original checkpoints and CEM budget.}
\label{tab:spatial}
\end{table}

\paragraph{A narrow ceiling.}
\Cref{tab:spatial} shows the result.
Selection rules from vision-language models fail in planning: ToMe and DART achieve only 0.51 and 0.34 of dense success, below random selection at 0.60; FastV and DivPrune barely exceed random selection.
Planning-aware selectors do better but stop in a narrow band.
The best, Guard-rank, reaches 0.89 of dense success, and CostGrad 0.84.
A further family fits this picture.
Probe-action responses improve on a fixed checkerboard by at most 1.2 points on paired episodes; shuffling the responses costs about 9 points.

\paragraph{Interpretation.}
Among planning-aware selectors the spread is small compared with their gap to the dense model, and on matched episodes a content-agnostic checkerboard is as good as response-based selection.
Saliency in the VLM sense, meaning tokens that attract attention or carry distinct content, is not what a dynamics predictor needs; uniform coverage of the grid is a better prior, a pattern that also holds for the context frame, as \cref{sec:context} shows.
As \cref{sec:axes} shows, the limiting factor is not \emph{which} tokens are kept but the fact that the same reduced token set is used at every step of every rollout, including the step that reads the real observation.
\textbf{Implication for SOAR:} which tokens are kept matters little, so SOAR keeps a fixed nested checkerboard and decides only how many tokens to compute.

\subsection{When matters more than which}
\label{sec:axes}

\begin{figure}[t]
\centering
\includegraphics[width=\linewidth]{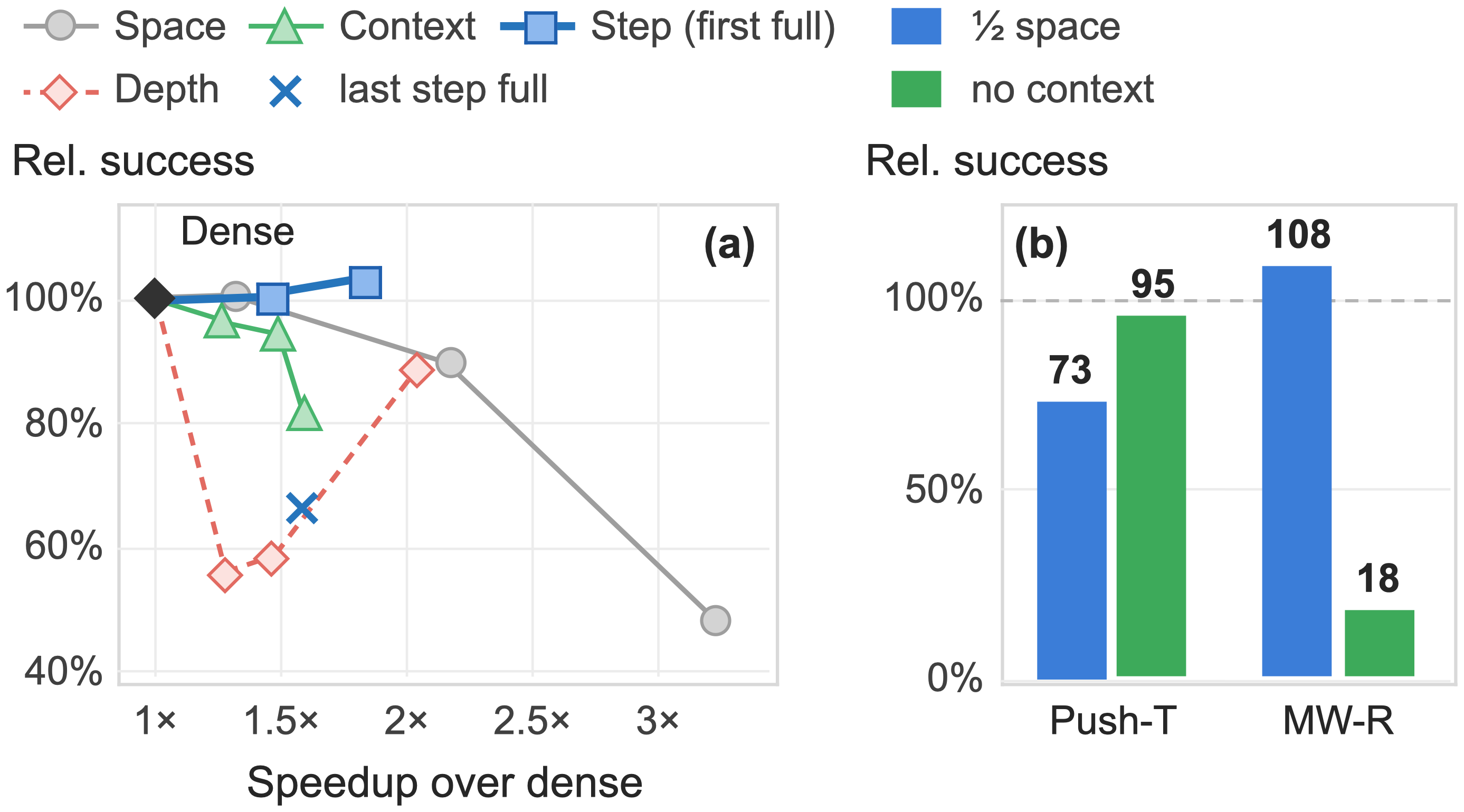}
\caption{\textbf{Where a planning rollout is redundant.} (a)~Success relative to dense versus speedup for the four pruning axes of \cref{tab:axes}. (b)~Success relative to dense on Push-T and MetaWorld Reach when halving the spatial positions or dropping the context frame.}
\label{fig:analysis}
\end{figure}

We now hold the selection rule fixed, using a nested checkerboard, and vary \emph{where} tokens are removed at comparable compute, as reported in \cref{tab:axes,fig:analysis}.
\textbf{Space} keeps 64, 128 or 192 positions in every frame of the rollout.
\textbf{Context frame} keeps the newest frame complete and keeps 0, 64 or 128 positions of the previous frame; with 0 the predictor sees a single frame.
\textbf{Rollout step} computes the first one or two steps in full and continues with 128 positions, or does the reverse and completes only the final step.
\textbf{Depth} uses all positions for the first $m$ layers and a checkerboard thereafter.

\begin{table}[t]
\centering
\small
\setlength{\tabcolsep}{3.5pt}
\begin{tabular}{llcccc}
\toprule
Axis & Configuration & Speedup & Avg. & Push-T & MW-R \\
\midrule
-- & Dense & 1.00$\times$ & 1.00 & 1.00 & 1.00 \\
\midrule
Space & 64 positions & 3.23$\times$ & 0.48 & 0.03 & 0.66 \\
 & 128 positions & 2.17$\times$ & 0.90 & 0.73 & 1.08 \\
 & 192 positions & 1.32$\times$ & 1.00 & 0.81 & 1.26 \\
\midrule
Context & 0 (single frame) & 1.59$\times$ & 0.82 & 0.95 & 0.18 \\
 & 64 positions & 1.49$\times$ & 0.95 & 0.90 & 1.32 \\
 & 128 positions & 1.27$\times$ & 0.97 & 0.95 & 1.18 \\
\midrule
Step & first step full & 1.82$\times$ & \textbf{1.03} & 0.74 & 1.13 \\
 & first two steps full & 1.45$\times$ & 1.01 & 0.77 & 1.34 \\
 & last step full & 1.59$\times$ & 0.66 & 0.16 & 1.18 \\
\midrule
Depth & 0 full layers & 2.04$\times$ & 0.89 & 0.73 & 1.11 \\
 & 1/3 full layers & 1.47$\times$ & 0.58 & 0.05 & 1.24 \\
 & 1/2 full layers & 1.28$\times$ & 0.56 & 0.00 & 1.08 \\
\bottomrule
\end{tabular}
\caption{\textbf{Pruning along four axes.} Speedup and success relative to dense: seven-task average, Push-T and MW-R.}
\label{tab:axes}
\end{table}

\paragraph{The first step anchors the ranking.}
Computing only step~0 in full and continuing with 128 positions raises success from 0.90 to 1.03 of dense at a $1.82\times$ speedup, matching the dense model, as shown in \cref{fig:analysis}a.
The gain holds in every evaluation, with an average margin of 5.4 points over the 128-position rollout, and the configuration retains 0.94 of dense success in \cref{tab:main}.
Placing comparable compute on the last step instead achieves only 0.66 of dense success.
The asymmetry follows from the structure of the rollout.
Step~0 is the only step that reads the real observation, and every later prediction, and hence the cost of every candidate, is conditioned on its output.
An error there is inherited by the entire rollout, whereas an error at a later step affects only what follows it.
Completing the last step is worse than pruning it, because the dense predictor sees an imputed state unlike anything seen in training.

\paragraph{Reduce the input, not the computation in flight.}
Processing all positions in the first layers and pruning afterwards is worse than pruning from the input: 0.58 and 0.56 versus 0.89 of dense, with Push-T falling to 5\% and 0\% of its dense successes.
We attribute this to a distribution shift: tokens that have attended to the full grid suddenly lose half of their neighbours, a configuration the dense-trained predictor never encounters.
FastV and DART, which keep the first two layers complete and prune afterwards, show the same pattern: they are slower than plain spatial pruning and no more accurate, as \cref{tab:spatial} shows; DART also fails completely on Push-T.
Together with the last-step result, this gives a simple rule for frozen predictors: a uniformly subsampled input is tolerated, but changing the token set inside a forward pass, or running the full model on an imputed state, is not.
The exact first step respects this rule: reduction happens only between steps and feeds the next step a cleanly subsampled input.

\paragraph{Tasks fail along different axes.}
Push-T and MW-R form the double dissociation shown in \cref{fig:analysis}b.
Halving the positions keeps 73\% of Push-T's dense successes but 108\% on MW-R; dropping the context frame keeps 95\% on Push-T but only 18\% on MW-R.
Pushing a block requires the spatial detail of the contact, whereas reaching with an arm requires information carried by the previous frame, as examined in \cref{sec:context}.
No single uniform budget serves both.
\textbf{Implication for SOAR:} the first step must stay exact, reductions must act on the input of a step, and the tolerable budget varies by task, so it must be chosen per decision.

\begin{figure*}[t]
\centering
\includegraphics[width=\textwidth]{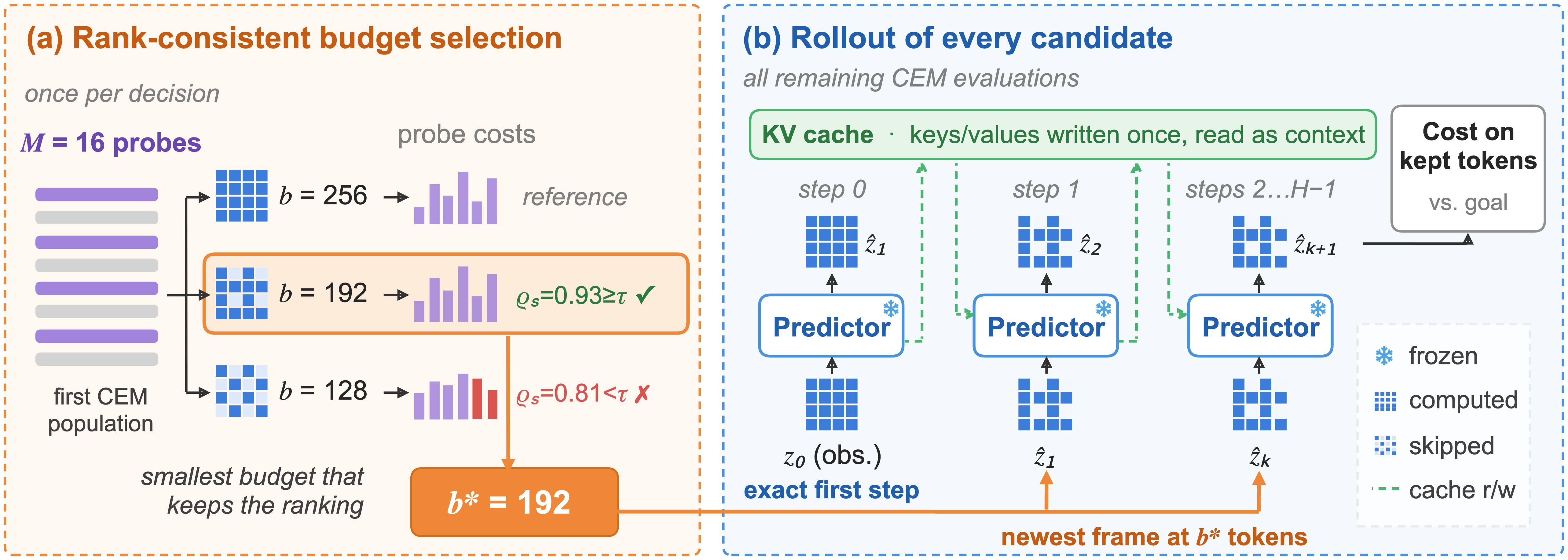}
\caption{\textbf{Overview of SOAR.} (a)~Once per decision, SOAR rolls out $M{=}16$ candidates of the first CEM population under three nested token budgets and selects the smallest budget $b^\star$ whose Spearman correlation with the full budget reaches $\tau$; the probe values are schematic. (b)~Every candidate rollout computes the first step in full, reads the context frame from a key--value cache instead of recomputing it, and computes the newest frame of later steps at $b^\star$ positions.}
\label{fig:method}
\end{figure*}

\subsection{The context frame must be kept, not recomputed}
\label{sec:context}

After step~0, half of every step recomputes the previous frame, whose outputs are discarded.
Dropping it is therefore the largest single saving available: the predictor then processes only the newest frame.
We measure its effect on every task; detailed results are in the supplementary material.

\paragraph{The value of context is a property of the task.}
On Maze, Wall, Push-T and Rc-Pl, the single-frame predictor matches the dense model, with mean differences between $-2.6$ and $+2.6$ points, so the planner does not use the information in the previous frame.
On MetaWorld the same change is catastrophic: MW-R loses 31 to 43 points and MW-RW 12 to 30.
Rc-R lies in between, with a mean loss of 6.8 points.
The split between the first four tasks and MetaWorld is consistent across our evaluations.
This is consistent with a single feature grid fixing the configuration of the scene but not its motion.
In navigation and pushing, the agent's next state is largely determined by the current configuration and the action; in arm reaching, the predictor needs two frames to infer the arm's current velocity.

\paragraph{How much context, not which context.}
When the context is needed, its value grows smoothly with the number of retained positions: success rises from 0.83 of dense without context to 0.95 with 64 checkerboard positions and 1.00 with the full frame.
Selecting positions by the magnitude of their predicted motion reaches only 0.92 of dense success versus 0.95 for a fixed checkerboard at 64 positions. A random subset of about 162 positions also performs comparably to a motion-selected subset of the same size, reaching 0.99 versus 1.01 of dense success.
Motion-selected positions cluster around moving objects and cover only 35--62\% of the $2\times2$ cells, whereas a checkerboard covers all of them.
The predictor reads the previous frame as a whole, and coverage matters more than concentration on motion.
This mirrors the saturation of spatial selection in \cref{sec:spatial}.

\paragraph{A trade-off that pruning cannot resolve.}
If the task type is known, using the single-frame predictor on quasi-static tasks and the first-step-complete rollout on arm tasks reaches 1.00 of dense success at a $1.89\times$ speedup.
This oracle shows the cost of the trade-off but requires a label that a planner does not have.
Pruning forces a choice between keeping the context, which pays for recomputation on every task, and dropping it, which fails on arm tasks.
Because the context frame was computed when it was newest, reading its cached keys and values keeps its information at the cost of a single-frame predictor.
\textbf{Implication for SOAR:} keep the context frame, but read it from a cache.

\section{SOAR: Rank-Consistent Compression}
\label{sec:method}

SOAR plans with a frozen latent world model at reduced cost by changing only how CEM evaluates candidate rollouts, as shown in \cref{fig:method}.
We first establish the criterion in \cref{sec:method_criterion}, reduce the design space to one scalar in \cref{sec:method_exact}, and select the budget online in \cref{sec:method_budget}.

\subsection{Rank consistency as the compression criterion}
\label{sec:method_criterion}
Let $c(a)$ be the cost of candidate $a$ under the dense rollout and $\tilde c(a)$ its cost under a compressed rollout.
CEM samples $N$ actions from $\mathcal{N}(\mu,\sigma^2)$, selects the $K$ lowest-cost indices as $\mathcal{E}$, and refits $(\mu,\sigma)$ to their actions.

\begin{proposition}
\label{prop:rank}
If $\tilde c = g\circ c$ for a strictly increasing function $g$, then, given the same random samples, CEM selects the same elite sets, follows the same sequence of sampling distributions and returns the same action under $\tilde c$ as under $c$.
\end{proposition}
\begin{proof}
A strictly increasing $g$ preserves the order of the $N$ costs, so the $K$ lowest-cost candidates, and hence $\mathcal{E}$, are unchanged.
The refit uses only the elite actions, not their costs, so $(\mu,\sigma)$ after the iteration is unchanged; induction over iterations gives the claim.
\end{proof}

The proposition has two consequences.
First, the error of a compressed rollout in feature or cost space is the wrong criterion: it penalizes compressions that change every cost but none of the decisions.
Second, what matters is membership in the top-$K$ set, which for $K{=}10$ of $N{=}300$ candidates requires near-perfect agreement at the top of the ranking.
The argument holds for any planner that uses costs only through their order, but not for cost-weighted schemes such as MPPI.

\subsection{Reducing the design space}
\label{sec:method_exact}
Measuring rank consistency online is affordable because the analysis of \cref{sec:analysis} fixes all choices but one.

\paragraph{Exact first step.}
Step~0 is the only step that reads the real observation, and every later prediction is conditioned on it.
Keeping all 256 positions recovers 5.4 points on average over compressing every step; see \cref{sec:axes}.

\paragraph{Cached context.}
From step~1 on, the predictor attends to the previous frame through the per-layer keys and values stored when that frame was the newest one, in the spirit of KV caching for autoregressive models~\cite{jha2026mstar}, so each frame is computed exactly once.
This keeps the context that arm tasks need, as \cref{sec:context} shows, at the cost of a single-frame predictor; from step~2 on, the cached keys and values carry one more frame of history than a recomputed context frame, as detailed in the supplement.

\paragraph{One knob remains.}
The only remaining decision is how many positions of the newest frame to compute at steps $1,\dots,H{-}1$.
SOAR uses one of three nested checkerboard sets, $\mathcal{B}=\{256,192,128\}$, because uniform coverage matters more than content-based selection, as shown in \cref{sec:spatial,sec:context}. It applies the reduction to each step's input so the predictor never switches token sets within a forward pass, consistent with \cref{sec:axes}.
For a horizon of 6, the three budgets cut compute by $1.8\times$, $2.3\times$ and $3.1\times$, close to the measured speedups of $1.8\times$, $2.2\times$ and $3.0\times$ reported in \cref{tab:soar}.

\begin{algorithm}[t]
\caption{SOAR at one planning decision}
\label{alg:soar}
\footnotesize
\begin{algorithmic}[1]
\Require observation $z_0$, goal $z_g$, CEM parameters $(N,K,T)$, budgets $\mathcal{B}$, probes $M$, threshold $\tau$
\State sample $a_{1:N}\sim\mathcal{N}(\mu,\sigma^2)$
\State $P\gets$ a fixed subset of $M$ candidates of $a_{1:N}$
\For{$b\in\mathcal{B}$}
  \State $\tilde c_b\gets\textsc{Rollout}(P;\,b)$ \Comment{exact step 0, cached context}
\EndFor
\State $b^\star \gets \min(\{256\} \cup
\{b \in \mathcal{B} :
\rho_s(\tilde{c}_b, \tilde{c}_{256}) \geq \tau\})$
\For{$t=1,\dots,T$}
  \State $c\gets\textsc{Rollout}(a_{1:N};\,b^\star)$
  \State $\mathcal{E}\gets$ indices of the $K$ lowest costs in $c$
  \State refit $(\mu,\sigma)$ to $\{a_i\}_{i\in\mathcal{E}}$; resample $a_{1:N}\sim\mathcal{N}(\mu,\sigma^2)$
\EndFor
\State \Return the first action(s) of $\mu$
\end{algorithmic}
\end{algorithm}

\subsection{Online budget selection}
\label{sec:method_budget}
The budget a decision tolerates differs across tasks, as shown in \cref{tab:axes}, and across decisions within a task, as shown in \cref{fig:teaser}. SOAR therefore estimates the budget at each decision using \cref{alg:soar}.
From the first CEM population it takes a fixed index subset of $M{=}16$ candidates, which costs no extra random numbers, rolls them out under every $b\in\mathcal{B}$, giving costs $\tilde c_b$, and chooses the budget:
\begin{equation}
b^\star = \min\left(
\{256\} \cup
\{b \in \mathcal{B} :
\rho_s(\tilde{c}_b, \tilde{c}_{256}) \geq \tau\}
\right).
\label{eq:budget}
\end{equation}

where $\rho_s$ is the Spearman rank correlation over the probe candidates and $\tau$ is a threshold shared across tasks.
Spearman correlation is a cheap proxy for the condition of \cref{prop:rank}: a value of one means the probes keep their order exactly.
The full budget serves as the fallback. The remaining CEM evaluations of the decision use $b^\star$, and decisions with a one-step remaining horizon skip the probe.

\paragraph{Cost.}
The probe adds at most $M|\mathcal{B}|{=}48$ rollouts per decision, two thirds of them at reduced budgets, against $N{=}300$ rollouts in each of 15--30 CEM iterations; this is 0.5--1.1\% of the rollouts of a decision, and its time is included in all reported speedups.
It draws no random numbers, so with $b^\star$ forced to 256, SOAR exactly reproduces the actions obtained with a fixed $b{=}256$. Holding $b^\star$ constant at 256, 192 or 128 gives the fixed-budget SOAR variants, isolating the exact first step and cache from budget selection.

\paragraph{Threshold.}
We fix $\tau{=}0.9$, and $0.8$ as a more aggressive setting, before any evaluation, from probe statistics on separate development episodes.
Rank agreement must be high, as \cref{prop:rank} suggests. On Push-T, \cref{tab:soar} reports 17 lost successes at 128 positions, despite a median $\rho_s$ of 0.85.

\section{Experiments}
\label{sec:experiments}

\begin{table}[t]
\centering
\small
\setlength{\tabcolsep}{3pt}
\begin{tabular}{lcc}
\toprule
Method & Success / Dense & Speedup \\
\midrule
\multicolumn{3}{l}{\emph{Spatial selection}} \\
CostGrad~\cite{xu2026costgrad} & 0.84 & 2.13$\times$ \\
Guard-rank & 0.89 & 1.92$\times$ \\
Checkerboard, 128 positions & 0.85 & 2.22$\times$ \\
\midrule
\multicolumn{3}{l}{\emph{Temporal structure}} \\
First step full & 0.94 & 1.89$\times$ \\
Single frame & 0.83 & 1.85$\times$ \\
Oracle task split & 1.00 & 1.89$\times$ \\
\midrule
\multicolumn{3}{l}{\emph{SOAR, our method}} \\
SOAR, $b{=}256$ & 1.02 & 1.76$\times$ \\
SOAR, $b{=}192$ & 1.01 & 2.21$\times$ \\
SOAR, $b{=}128$ & 0.93 & 2.98$\times$ \\
Adaptive SOAR, $\tau{=}0.9$ & 1.01 & 2.31$\times$ \\
Adaptive SOAR, $\tau{=}0.8$ & 0.98 & 2.54$\times$ \\
\midrule
\multicolumn{3}{l}{\emph{Trained sparse world model}} \\
Scope-WM~\cite{li2026scopewm} & 1.02 & 2.04$\times$ \\
\bottomrule
\end{tabular}
\caption{\textbf{Speed--quality comparison.} Success relative to dense and planning speedup on paired batches.}
\label{tab:main}
\end{table}

\begin{table*}[t]
\centering
\small
\setlength{\tabcolsep}{8pt}
\begin{tabular}{lccccccccc}
\toprule
Method & Maze & Wall & Push-T & MW-R & MW-RW & Rc-R & Rc-Pl & Avg. & Speedup \\
\midrule
Dense & 72 & 76 & 60 & 47 & 32 & 2 & 12 & 48.96 & 1.00$\times$ \\
SOAR, $b{=}256$ & 75 & 77 & 64 & 54 & 25 & 4 & 10 & 50.15 & 1.76$\times$ \\
SOAR, $b{=}192$ & 76 & 68 & 49 & 56 & 28 & 5 & 13 & 49.26 & 2.21$\times$ \\
SOAR, $b{=}128$ & 61 & 72 & 43 & 50 & 28 & 6 & 11 & 45.39 & 2.98$\times$ \\
\midrule
SOAR, $\tau{=}0.9$ & 75 & 74 & 56 & 52 & 23 & 6 & 11 & \textbf{49.26} & \textbf{2.31$\times$} \\
SOAR, $\tau{=}0.8$ & 74 & 70 & 49 & 46 & 34 & 6 & 11 & 48.21 & 2.54$\times$ \\
\bottomrule
\end{tabular}
\caption{\textbf{SOAR versus fixed budgets.} Successes out of 96 episodes per task, except 32 on RoboCasa; average success in percent and speedup over dense. SOAR matches the 192-position budget on average at a higher speedup while avoiding its losses on Push-T and Wall.}
\label{tab:soar}
\end{table*}

\paragraph{Temporal structure moves the frontier.}
\Cref{tab:main} compares all methods against the dense model of the same batch.
Spatial selectors sit at 0.84--0.89 of dense success at about a $2\times$ speedup.
Keeping the first step complete reaches 0.94 at a $1.89\times$ speedup, and reusing the cached context reaches 1.02 of dense success at a $1.76\times$ speedup without knowing the task, so the extra frame of history carried by cached keys and values does not hurt planning, as explained in \cref{sec:method_exact}.
With the newest frame at 192 or 128 positions, the fixed-budget variants reach $2.2\times$ and $3.0\times$ speedups at 1.01 and 0.93 of dense success, on or above the frontier of spatial selection and comparable to Scope-WM~\cite{li2026scopewm}, which reaches dense-level success at $2.04\times$ only after training a sparse model in our five-task reimplementation.

\paragraph{Adaptive budget.}
\Cref{tab:soar} compares SOAR with its fixed-budget variants on the same paired episodes.
Before the run, we registered the thresholds $\tau\in\{0.9,0.8\}$ and three success criteria: a speedup of at least $2\times$, success within one point of SOAR with $b{=}256$, and Push-T within five points of dense.
SOAR with $\tau{=}0.9$ meets all three.
It reaches 1.01 of dense success at a $2.31\times$ speedup; the difference is $+0.3$ points with a 95\% confidence interval of $[-4.2,+4.9]$.

\paragraph{Where adaptivity helps.}
SOAR matches the average of the fixed budget closest in speed, 192 positions, which could only be chosen in hindsight by a closed-loop sweep, but with a different failure profile.
Relative to dense, the fixed budget loses 11 Push-T successes and 8 on Wall, the tasks that \cref{tab:axes} identifies as sensitive to spatial resolution; SOAR loses only 4 and 2.
It gives up four MW-R successes and five MW-RW successes relative to the fixed budget.
On MW-RW, per-task variation is large across all variants: the fixed 256-position budget scores 25 and $\tau{=}0.8$ scores 34, against 32 for dense.
SOAR thus trades a few successes on compression-tolerant tasks for robustness on the others.

\balance 
\paragraph{The probe recovers the task structure.}
The selected budgets shown on the right of \cref{fig:teaser} reproduce the dissociation of \cref{sec:axes} without any task label.
On MetaWorld and RoboCasa, SOAR uses 128 positions on 61--100\% of decisions; on Maze it keeps the full grid on 97\% of decisions, and its $1.82\times$ speedup nearly matches the $1.85\times$ of SOAR with $b{=}256$, which shows that the probe itself is nearly free.
Push-T and Wall use all three budgets, showing adaptation across decisions rather than just across tasks.

\paragraph{Where the speedup comes from.}
The speedup of SOAR follows the selected budgets: $1.8\times$ on Maze, where it keeps the full grid, $2.2\times$ on Wall, $2.3\times$ on Push-T and $2.6$--$2.7\times$ on MetaWorld, where 128 positions dominate; on RoboCasa, where two of three steps are compressible, $2.3$--$2.4\times$.

\paragraph{The probe measures what matters.}
On the development episodes used to fix $\tau$, where the planner always uses the full budget, the median $\rho_s$ at 192 positions is 0.84--0.91 on Maze, Push-T and Wall and 0.95--0.99 on MetaWorld and RoboCasa.
This ordering matches independent closed-loop evaluation: \cref{tab:soar} reports that the fixed 128-position budget loses 11 and 17 successes on Maze and Push-T but no more than four on any other evaluated task.

\paragraph{Threshold.}
Lowering $\tau$ to 0.8 raises speedup to $2.54\times$ but sends 77\% of Push-T decisions to 128 positions; Push-T then loses 11 successes against dense, as with the fixed 192-position budget, rather than four at $\tau{=}0.9$.

\paragraph{Exactness checks.}
We test the implementation on small randomly initialized predictors with the released architectures.
Recomputing the context frame at every step reproduces the original sliding-window rollout exactly, and reading it from the cache reproduces a single forward pass of the same predictor over the imagined trajectory with a banded causal mask.
With $\tau{>}1$ or $\tau{<}{-1}$, SOAR exactly recovers the actions at $b{=}256$ or $b{=}128$, respectively; model weights, goal encoding and CEM randomness remain unchanged.

\section{Limitations and Conclusion}
\label{sec:conclusion}

\paragraph{Limitations.}
Our analysis uses the released JEPA-WM checkpoints, which share a two-frame predictor, and \cref{prop:rank} covers planners that select candidates by rank; extending the study to longer-context predictors and cost-weighted planners such as MPPI is future work.
The probe estimates rank consistency on first-population candidates, which are more diverse than later elites; probing later iterations may compress further.
Further extensions are a top-$K$ overlap criterion for the probe and finer budget sets.

\paragraph{Conclusion.}
Token reduction for world-model planning has followed the spatial recipe of vision-language models, yet a planner needs only the candidate ranking, and what matters is when computation is spent, not which tokens are kept.
With frozen models, SOAR keeps the first step exact, computes each frame once, and uses candidate rankings to achieve dense-level planning at a $2.3\times$ speedup.

\clearpage
\nobalance 
\section*{LLM Usage Statement}
We used Claude (Anthropic) to polish the paper's wording. The experiments and data are the authors' own; the model did not design or run experiments or change results. The authors reviewed all text and figures and take full responsibility for the paper.
{
    \small
    \bibliographystyle{ieeenat_fullname}
    \bibliography{main}
}

\clearpage
\setcounter{section}{0}
\setcounter{table}{0}
\setcounter{figure}{0}
\setcounter{equation}{0}
\renewcommand{\thesection}{\Alph{section}}
\renewcommand{\theHsection}{supp.\Alph{section}}
\renewcommand{\theHsubsection}{supp.\Alph{section}.\arabic{subsection}}
\renewcommand{\theHsubsubsection}{supp.\Alph{section}.\arabic{subsection}.\arabic{subsubsection}}
\renewcommand{\theHtable}{supp.\arabic{table}}
\renewcommand{\theHfigure}{supp.\arabic{figure}}
\renewcommand{\theHequation}{supp.\arabic{equation}}
\renewcommand{\thetable}{S\arabic{table}}
\renewcommand{\thefigure}{S\arabic{figure}}
\renewcommand{\theequation}{S\arabic{equation}}
\maketitlesupplementary

\section{Evaluation and Implementation Details}
\label{sec:supp_protocol}
This supplement provides the evaluation settings, full spatial-baseline results, temporal ablations, details of the cached rollout and ranking probe, additional SOAR analyses, trained-model controls, and alternative acceleration experiments. References to numbered sections and tables in the main paper are explicitly marked as such; all tables here use an S prefix.

\subsection{Models, tasks, and planning budget}
We use released JEPA-WM checkpoints~\cite{terver2025jepawms}. The visual encoder is frozen and runs on the complete image in all experiments. Each frame has $16\times16=256$ patch positions. The predictor is an action-conditioned AdaLN transformer with six layers and width 384 for Maze, Wall, Push-T and MetaWorld, and twelve layers and width 1024 for RoboCasa. The corresponding encoders are DINOv2 ViT-S/14 and DINOv3 ViT-L/16.

CEM samples 300 candidates, selects ten elites, and retains the released search budget. Maze, Wall and Push-T use 30 search iterations, a horizon of six, and execute six actions per decision. MetaWorld uses 15 iterations, a horizon of six, and executes three actions per decision. RoboCasa uses 15 iterations, a horizon of three, and executes one action per decision. An episode has one planning call on the first three tasks, seven on MetaWorld and twelve on RoboCasa. The remaining horizon can shrink near the end of an episode. The objective uses the final predicted frame's L2 distance to the encoded goal, with a $0.1$-weighted proprioceptive term where applicable.

Every complete evaluation contains 96 episodes per task on the first five tasks and 32 per RoboCasa task, or 544 episodes across seven tasks for one evaluation seed. Three such seeds give 1,632 episodes. Development panels and offline probe diagnostics are reported separately from complete closed-loop evaluations. 

\subsection{Aggregation and uncertainty}
Let $y_{m,t,s,i}\in\{0,1\}$ be the outcome for method $m$, task $t$, evaluation seed $s$ and episode $i$. The task success rate and macro-average are
\begin{equation}
 S_{m,t}=\frac{1}{|\mathcal S_t|}\sum_s\frac{1}{n_{t,s}}\sum_i y_{m,t,s,i},\qquad
 \bar S_m=\frac{1}{|\mathcal T|}\sum_t S_{m,t}.
\end{equation}
Success relative to dense is $\bar S_m/\bar S_D$, using the dense control for the corresponding evaluation. A difference in percentage points is $100(\bar S_m-\bar S_D)$. Task macro-averages weight all seven tasks equally despite their different episode counts.

Methods share initial states and goals within a comparison. The main paper's confidence intervals use 10,000 paired bootstrap resamples stratified by task, preserving the method--control pairing. For tables with seed SDs, captions specify the SD convention and whether the seeds refer to training or evaluation.

\subsection{Timing and memory}
Evaluations use H100 GPUs and FP32 without autocast. The timing scope is one complete \texttt{GC\_Agent.act} call: observation encoding, token selection or ranking probes, CEM search and all auxiliary predictor evaluations. Simulator execution, goal pre-encoding and file I/O are excluded. SOAR timing excludes the first five decisions of each job as warm-up. Latency is the mean wall-clock time per planning decision.

For each task, let $L_{m,t}$ denote the complete planning latency. The main speed metric uses $r_m=|\mathcal T|^{-1}\sum_t L_{m,t}/L_{D,t}$ and reports $1/r_m$. This task-normalized metric weights tasks equally; $\sum_t L_{D,t}/\sum_t L_{m,t}$ instead weights by dense latency. Summaries that use other timing weights specify them in their captions. Spatial budgets denote retained position counts, and complete planning latency is measured separately.

Timing summaries average all recorded runs within each experiment family and use the dense control from that family. \Cref{tab:supp_memory} reports peak PyTorch allocated memory.

\section{Spatial Baselines and Recovery Interfaces}
\label{sec:supp_spatial}
\begin{table*}[!t]
\centering
\small
\setlength{\tabcolsep}{3.5pt}
\begin{tabular}{lcccccccc}
\toprule
Method & Maze & Wall & Push-T & MW-R & MW-RW & Rc-R & Rc-Pl & Avg. \\
\midrule
{Dense} & {79.51\,\scriptsize(2.41)} & {77.08\,\scriptsize(1.80)} & {65.28\,\scriptsize(1.20)} & {48.26\,\scriptsize(4.34)} & {23.61\,\scriptsize(4.21)} & {16.67\,\scriptsize(4.77)} & {35.42\,\scriptsize(15.42)} & {49.40} \\
{Random 50\%} & {64.93\,\scriptsize(3.35)} & {34.72\,\scriptsize(3.66)} & {11.81\,\scriptsize(1.59)} & {35.76\,\scriptsize(0.60)} & {10.42\,\scriptsize(3.12)} & {12.50\,\scriptsize(5.41)} & {38.54\,\scriptsize(13.01)} & {29.81} \\
{ToMe 50\%} & {49.31\,\scriptsize(9.85)} & {34.72\,\scriptsize(1.59)} & {11.46\,\scriptsize(3.76)} & {20.14\,\scriptsize(4.21)} & {8.33\,\scriptsize(3.61)} & {12.50\,\scriptsize(5.41)} & {38.54\,\scriptsize(15.73)} & {25.00} \\
{FastV} & {31.60\,\scriptsize(6.94)} & {26.39\,\scriptsize(2.62)} & {55.56\,\scriptsize(1.59)} & {35.42\,\scriptsize(3.76)} & {23.61\,\scriptsize(4.21)} & {18.75\,\scriptsize(10.83)} & {38.54\,\scriptsize(4.77)} & {32.84} \\
{DART} & {31.25\,\scriptsize(4.54)} & {15.28\,\scriptsize(1.59)} & {0.00\,\scriptsize(0.00)} & {25.35\,\scriptsize(4.21)} & {5.56\,\scriptsize(0.60)} & {21.88\,\scriptsize(3.12)} & {17.71\,\scriptsize(10.05)} & {16.72} \\
{DivPrune 50\%} & {65.97\,\scriptsize(1.20)} & {38.54\,\scriptsize(6.25)} & {34.38\,\scriptsize(1.04)} & {30.21\,\scriptsize(1.80)} & {7.29\,\scriptsize(1.04)} & {7.29\,\scriptsize(3.61)} & {35.42\,\scriptsize(6.51)} & {31.30} \\
{CostGrad 50\%} & {76.74\,\scriptsize(6.78)} & {51.74\,\scriptsize(4.21)} & {61.46\,\scriptsize(3.76)} & {36.46\,\scriptsize(8.33)} & {3.82\,\scriptsize(2.41)} & {15.62\,\scriptsize(3.12)} & {44.79\,\scriptsize(6.51)} & {41.52} \\
\bottomrule
\end{tabular}
\caption{\textbf{Spatial baselines: success rate (\%).} Means and sample SDs over evaluation seeds 1, 101 and 1001; 96 episodes per seed on the first five tasks and 32 on each RoboCasa task. Avg. weights tasks equally. CostGrad here uses the retained-position goal cost; CostGrad-copy in \cref{tab:supp_recovery} uses the full-grid cost.}
\label{tab:supp_spatial}
\end{table*}
\begin{table*}[!t]
\centering
\small
\setlength{\tabcolsep}{3.5pt}
\begin{tabular}{lccccccc}
\toprule
Method & Maze & Wall & Push-T & MW-R & MW-RW & Rc-R & Rc-Pl \\
\midrule
Dense & 45.864 & 45.922 & 45.847 & 19.933 & 19.895 & 50.540 & 50.625 \\
Random 50\% & 20.642 & 20.633 & 20.584 & 8.974 & 8.978 & 25.770 & 25.898 \\
ToMe 50\% & 20.723 & 20.686 & 20.685 & 9.003 & 9.035 & 25.870 & 25.972 \\
FastV & 48.418 & 33.087 & 33.117 & 14.405 & 14.393 & 31.106 & 31.165 \\
DART & 31.184 & 31.146 & 31.058 & 13.556 & 13.503 & 30.935 & 30.927 \\
DivPrune 50\% & 20.554 & 20.609 & 20.561 & 8.984 & 9.014 & 25.848 & 37.272 \\
CostGrad 50\% & 20.640 & 20.644 & 20.606 & 8.999 & 9.050 & 25.910 & 25.834 \\
\bottomrule
\end{tabular}
\caption{\textbf{Spatial baselines: complete planning latency (seconds/decision).} Mean complete planning time per decision on each task, with the original CEM search budget and the timing scope defined in Sec.~A.3.}
\label{tab:supp_latency}
\end{table*}
\begin{table*}[!t]
\centering
\small
\setlength{\tabcolsep}{3.5pt}
\begin{tabular}{lccccccc}
\toprule
Method & Maze & Wall & Push-T & MW-R & MW-RW & Rc-R & Rc-Pl \\
\midrule
Dense & 15.67 & 15.67 & 15.67 & 15.71 & 15.71 & 14.74 & 14.74 \\
Random 50\% & 5.74 & 5.74 & 5.74 & 5.78 & 5.78 & 9.46 & 9.46 \\
ToMe 50\% & 5.86 & 5.86 & 5.86 & 5.90 & 5.90 & 9.75 & 9.75 \\
FastV & 15.45 & 15.45 & 15.45 & 15.49 & 15.49 & 16.49 & 16.49 \\
DART & 15.45 & 15.45 & 15.45 & 15.49 & 15.49 & 16.49 & 16.49 \\
DivPrune 50\% & 5.74 & 5.74 & 5.74 & 5.78 & 5.78 & 9.46 & 9.46 \\
CostGrad 50\% & 5.78 & 5.78 & 5.78 & 5.81 & 5.81 & 9.49 & 9.49 \\
\bottomrule
\end{tabular}
\caption{\textbf{Peak PyTorch allocated memory (GiB).} For ToMe, FastV and DART, the peak statistic includes warm-up calls.}
\label{tab:supp_memory}
\end{table*}
\subsection{How the methods are adapted}
Random samples 128 distinct positions without replacement. DivPrune~\cite{alvar2025divprune} uses greedy diversity selection from cosine distances between DINO features. CostGrad~\cite{xu2026costgrad} performs a full one-step probe, differentiates its goal error with respect to visual input tokens, and retains the 128 largest gradient norms. Only input gradients are computed; predictor weights are not updated. Its mask stays fixed across candidates, CEM iterations and rollout steps of a real decision and is reconsidered when a new observation arrives.

ToMe~\cite{bolya2023tome} performs bipartite matching at the predictor entrance, merges to 128 groups, uses proportional attention, and reconstructs through a delta update. FastV~\cite{chen2024fastv} runs the first two layers on the full grid, then uses actual attention scores to choose 128 positions on each forward pass. DART~\cite{wen2025dart} also computes two full layers before selecting visual pivots and removing redundant features. These methods compute full-grid layers before pruning, adding work relative to pruning at the predictor entrance.

\subsection{State reconstruction and goal cost}
The spatial adaptations in \cref{tab:supp_spatial} use method-specific reconstruction. For Random and DivPrune, deleted positions retain the preceding frame's features and the goal cost uses the complete grid. CostGrad uses the retained-position cost. The compact rollout used for the temporal-axis study propagates only retained positions and restricts the encoded goal to those positions.

\Cref{tab:supp_recovery} reports an additional complete evaluation of the reconstruction interface. CostGrad-copy restores deleted positions with previous features and evaluates the full-grid goal cost. Its MW-R and Rc-R means are 36.11\% and 16.67\%, compared with 36.46\% and 15.62\% for retained-cost CostGrad.

Guard-rank selects a token-selection/reconstruction recipe at each real planning decision. It evaluates 17 fixed action probes and compares the action rankings from six recipes with the dense ranking to choose the recipe. Grid-guide retains a fixed checkerboard, evaluates nine full probes, and uses them to choose a reference trajectory for correcting candidate predictions. Guard-rank gains 2.33 macro-average points over CostGrad-copy. In the recovery controls' five-task normalized timing summary, planning latency is 12.574 seconds/decision for Guard-rank and 10.963 for CostGrad-copy, an increase of 14.7\%. Its Push-T success falls by 9.38 points, and success also decreases on both RoboCasa tasks.

\begin{table*}[!t]
\centering
\small
\setlength{\tabcolsep}{3.5pt}
\begin{tabular}{lcccccccc}
\toprule
Method & Maze & Wall & Push-T & MW-R & MW-RW & Rc-R & Rc-Pl & Avg. \\
\midrule
{CostGrad-copy} & {76.74\,\scriptsize(6.78)} & {51.74\,\scriptsize(4.21)} & {61.46\,\scriptsize(3.76)} & {36.11\,\scriptsize(8.36)} & {3.82\,\scriptsize(2.41)} & {16.67\,\scriptsize(1.80)} & {44.79\,\scriptsize(6.51)} & {41.62} \\
{Guard-rank} & {76.39\,\scriptsize(5.14)} & {63.19\,\scriptsize(3.35)} & {52.08\,\scriptsize(3.76)} & {52.78\,\scriptsize(2.62)} & {14.24\,\scriptsize(2.62)} & {11.46\,\scriptsize(7.86)} & {37.50\,\scriptsize(3.12)} & {43.95} \\
{Grid-guide} & {79.17\,\scriptsize(1.80)} & {66.32\,\scriptsize(5.74)} & {38.19\,\scriptsize(5.74)} & {47.22\,\scriptsize(3.01)} & {15.62\,\scriptsize(2.76)} & {14.58\,\scriptsize(6.51)} & {34.38\,\scriptsize(10.83)} & {42.21} \\
\bottomrule
\end{tabular}
\caption{\textbf{Recovery-interface controls: success (\%).} Same evaluation seeds and episode counts as \cref{tab:supp_spatial}. Parentheses denote sample SD over evaluation seeds. CostGrad-copy uses the full-grid goal cost.}
\label{tab:supp_recovery}
\end{table*}

\paragraph{Per-task variation and timing outliers.}
DART records no Push-T successes, while FastV retains much more of the dense Push-T performance despite its lower overall mean. Rc-Pl exhibits substantial variation across seeds, as reflected in the SDs in \cref{tab:supp_spatial}. The FastV/Maze run at seed 101 takes 78.905 seconds per decision, contributing to the 48.418-second task mean. DivPrune/Rc-Pl at seed 1001 takes 59.971 seconds, contributing to its 37.272-second task mean. The sources of these timing outliers have not been identified.

\section{Temporal Ablations and Context Coverage}
\label{sec:supp_context}
The four-axis study fixes the spatial selection rule and changes where computation is reduced. Spatial variants retain 64, 128 or 192 positions at every step. Context variants retain the full newest frame and 0, 64 or 128 positions of its predecessor. Step variants compute the first step, first two steps, or last step in full. Depth variants compute a full prefix of predictor layers and reduce the positions thereafter. These configurations retain the original planner budget.

\subsection{First-step anchoring across evaluations}
\Cref{tab:supp_seed_macro} expands the repeated temporal evaluations. For the original four matched evaluation seeds (5001, 6001, 8001 and 9001), anchoring improves macro success in every case relative to the 128-position rollout. The mean difference from the rounded entries is 5.43 points. The task split uses a single frame for Maze, Wall, Push-T and Rc-Pl, and the first-step-full two-frame rollout for MW-R, MW-RW and Rc-R. This task-informed split was determined using seed 5001; its entry is marked as in-sample.

\begin{table}[!htbp]
\centering\small\setlength{\tabcolsep}{3pt}
\begin{tabular}{lrrrrr}
\toprule
Seed & Dense & Spatial 128 & First full & Single & Split \\
\midrule
5001 & 46.73 & 41.82 & 48.21 & 38.24 & 48.81$^\dagger$ \\
6001 & 54.02 & 42.11 & 51.04 & 44.94 & 55.80 \\ 8001 & 42.71 & 35.57 & 39.58 & 33.04 & 40.48 \\
9001 & 49.70 & 44.05 & 46.43 & 42.56 & 50.00 \\

\bottomrule
\end{tabular}
\caption{\textbf{Temporal configurations in the original four matched evaluations.} Seven-task macro success (\%). Single drops the previous frame. $^\dagger$The split is selected using seed 5001, so this entry is in-sample. These four seeds underlie the first-step anchoring comparison in the main paper.}
\label{tab:supp_seed_macro}
\end{table}

\subsection{Dropping the previous frame}
\Cref{tab:supp_context_seeds} reports the per-task seed breakdown. Removing the previous frame reduces success on both MetaWorld tasks in every evaluation. Navigation and pushing have smaller average changes, with variation across seeds: Maze at seed 8001 loses 14.6 points.

\begin{table*}[!t]
\centering\small\setlength{\tabcolsep}{6pt}
\begin{tabular}{lrrrrrrr}
\toprule
Seed & Maze & Wall & Push-T & MW-R & MW-RW & Rc-R & Rc-Pl \\
\midrule
5001 & 0.0 & $-$3.1 & $-$3.1 & $-$32.3 & $-$17.7 & $-$6.2 & 3.1 \\
6001 & 1.0 & 1.0 & 2.1 & $-$35.4 & $-$13.5 & $-$18.8 & 0.0 \\
8001 & $-$14.6 & $-$1.0 & 1.0 & $-$42.7 & $-$16.7 & 3.1 & 3.1 \\
9001 & 0.0 & $-$3.1 & $-$3.1 & $-$31.2 & $-$12.5 & $-$3.1 & 3.1 \\
10001 & $-$4.2 & 4.2 & 1.0 & $-$34.4 & $-$19.8 & $-$18.8 & 0.0 \\
11001 & 2.1 & 5.2 & 5.2 & $-$34.4 & $-$30.2 & 3.1 & 6.2 \\
\midrule
Mean & $-$2.6 & 0.5 & 0.5 & $-$35.1 & $-$18.4 & $-$6.8 & 2.6 \\
\bottomrule
\end{tabular}
\caption{\textbf{Effect of dropping the context frame.} Single-frame success minus dense success, in percentage points. Each seed has 96 episodes per task on the first five tasks and 32 on each RoboCasa task. Differences are rounded to one decimal place.}
\label{tab:supp_context_seeds}
\end{table*}

\subsection{Context amount versus context selection}
With the newest frame complete, checkerboard context selection achieves 46.4\% macro success at 64 positions, versus 45.1\% for motion selection. At about 162 positions, random and motion-selected subsets achieve 48.6\% and 49.4\%, respectively (\cref{tab:supp_context_budget}). Motion-selected positions cover 35--62\% of the $2\times2$ grid cells, while the checkerboard covers every cell. These observations support spatial coverage as a useful prior for context selection.

\begin{table}[!htbp]
\centering\small\setlength{\tabcolsep}{4pt}
\begin{tabular}{lrrr}
\toprule
Context rule & Positions & Speedup & Success \\
\midrule
Dropped & 0 & 1.92$\times$ & 40.8 \\
Checkerboard & 64 & 1.56$\times$ & 46.4 \\
Top motion & 64 & 1.59$\times$ & 45.1 \\
Random & $\approx$162 & 1.19$\times$ & 48.6 \\
Motion, 90\% energy & $\approx$162 & 1.19$\times$ & 49.4 \\
Complete (dense) & 256 & 1.00$\times$ & 49.0 \\
\bottomrule
\end{tabular}
\caption{\textbf{Context amount and selection.} Macro success (\%) over two evaluation seeds, or 1,088 episodes per row. Only the previous frame is reduced. Speedup is the inverse task-normalized latency ratio.}
\label{tab:supp_context_budget}
\end{table}

\section{Cached Rollouts and Rank Consistency}
\label{sec:supp_kvc}
\subsection{Cached rollout reference}
KV caching is an existing autoregressive inference technique~\cite{jha2026mstar}. The first prediction uses the complete observed feature grid. In the uncompressed cached variant, the next step is identical to the original rollout: the cached observation was computed alone at step~0, as it would be when recomputed as the oldest frame of the two-frame window. From step~2 onward, the cached representations were computed while attending to their own predecessor. They therefore contain an additional historical dependency compared with recomputing that frame in isolation as the oldest frame of a fresh window.

The cached rollout corresponds to a single forward pass of the same predictor over the imagined trajectory with a banded frame-causal attention mask. Each frame attends to itself and its immediate predecessor, and information can propagate through layers. Cached keys and values at each layer match this reference. Rotary position encodings use absolute temporal indices. The current--previous temporal displacement remains one, as in the original window, because the corresponding rotary attention scores depend on index differences.

The full-budget SOAR variant isolates the first-step-full cached rollout, while the lower budgets additionally compress the newest-frame computation. Budget selection is held fixed within a real planning decision. The final cost compares predicted state and goal features at the same retained spatial indices.

\subsection{Token-computation accounting}
For $H$ predicted steps and $L$ predictor layers, dense visual token--layer work is proportional to
\begin{equation}
 C_D=L\,[256+512(H-1)].
\end{equation}
With a complete first step and a cached predecessor, computing $b$ newest-frame positions at each remaining step gives
\begin{equation}
 C_b=L\,[256+b(H-1)].
\end{equation}
For $H=6$, the ratios $C_b/C_D$ at $b=256,192,128$ are 0.545, 0.432 and 0.318. For $H=3$ they are 0.600, 0.500 and 0.400. These ratios account for visual token--layer work. Complete planning latency additionally includes probes, encoding and CEM updates, as well as kernel overhead and attention over cached keys.

\subsection{Ranking criterion and CEM updates}
The proposition in main-paper Sec.~5.1 relates candidate ordering to CEM updates. At an iteration, let $\mathcal E_D$ and $\mathcal E_b$ be the ten lowest-cost candidates under the dense and reduced costs. If $\mathcal E_D=\mathcal E_b$ and refitting uses only their actions, the next CEM distribution is unchanged. With the same random draws, this equality inductively preserves the complete search. A strictly increasing transform of all candidate costs is one sufficient way to ensure it. Consistent tie handling is required when costs tie.

SOAR estimates rank consistency from 16 candidates in the first 300-candidate CEM population. It selects the smallest tested budget whose probe-cost Spearman correlation with the full-budget cached rollout reaches the shared threshold. CEM then uses that budget to evaluate its candidates and refit the search distribution.

\subsection{Probe implementation}
The 16 probe indices are deterministic and evenly spaced in the initial population:
\begin{equation}
 p_i=\operatorname{round}(299i/15),\quad i=0,\ldots,15.
\end{equation}
The 16 action sequences are evaluated at nested checkerboard budgets of 128, 192 and 256 positions. Nonconstant probe-cost vectors use Spearman correlation with average ranks for ties; the constant-input branch assigns a selection score of one, which passes both thresholds. Budgets are tested in order, 128 then 192, with 256 as fallback. The budget stays fixed for the remaining CEM evaluations of that decision. One-step decisions use the complete first step and skip the probe.

The full-budget \emph{cached} rollout supplies the reference costs. For each tested budget, probe goal features use the matching final spatial positions. The probe reuses candidates already sampled for the first CEM population.

The probe adds at most 48 rollouts per decision, two-thirds at reduced budgets. This is 1.07\% of the $300\times15$ or 0.53\% of the $300\times30$ candidate rollouts in the main search.

\paragraph{Implementation checks.}
The 261 checks use small randomly initialized predictors with the released architectures, including twelve-layer RoboCasa. They verify recomputed context against the original rollout, caches against an independent banded-mask reference, unchanged weights and CEM random-number state, and one-step horizons. Forcing 256 or 128 positions reproduces fixed-budget actions. Decision audits recompute ranks and budgets from logged costs and check indices, forward counts, rollout layouts and final positions.

\twocolumn[{
\section{SOAR: Budget Allocation and Success Profiles}
\label{sec:supp_soar}
We expand main-paper Table~4 with per-task latency ratios, budget allocations and probe diagnostics.

\noindent\begin{minipage}{\textwidth}
\centering\small\setlength{\tabcolsep}{4pt}
\begin{tabular}{lcccccccr}
\toprule
Method & Maze & Wall & Push-T & MW-R & MW-RW & Rc-R & Rc-Pl & Avg. \\
\midrule
Dense & 72 (1.00) & 76 (1.00) & 60 (1.00) & 47 (1.00) & 32 (1.00) & 2 (1.00) & 12 (1.00) & 48.96 \\
Fixed 256 & 75 (0.54) & 77 (0.54) & 64 (0.54) & 54 (0.55) & 25 (0.55) & 4 (0.63) & 10 (0.62) & 50.15 \\
Fixed 192 & 76 (0.42) & 68 (0.42) & 49 (0.42) & 56 (0.43) & 28 (0.43) & 5 (0.52) & 13 (0.52) & 49.26 \\
Fixed 128 & 61 (0.30) & 72 (0.30) & 43 (0.30) & 50 (0.31) & 28 (0.31) & 6 (0.42) & 11 (0.41) & 45.39 \\
SOAR, $\tau=0.9$ & 75 (0.55) & 74 (0.45) & 56 (0.43) & 52 (0.37) & 23 (0.39) & 6 (0.43) & 11 (0.42) & 49.26 \\
SOAR, $\tau=0.8$ & 74 (0.51) & 70 (0.36) & 49 (0.35) & 46 (0.34) & 34 (0.35) & 6 (0.43) & 11 (0.42) & 48.21 \\
\bottomrule
\end{tabular}
\captionof{table}{\textbf{SOAR success counts and latency profiles.} Entries are successes out of 96 episodes (32 on RoboCasa); parentheses give complete planning latency divided by dense on that task, rounded to two decimals. Avg. is task-macro success (\%). Overall speedups use unrounded normalized latencies, not these rounded entries.}
\label{tab:supp_soar_detail}
\end{minipage}\par\vspace{8pt}

\noindent\begin{minipage}{\textwidth}
\centering\small\setlength{\tabcolsep}{6pt}
\begin{tabular}{lccccccc}
\toprule
Threshold & Maze & Wall & Push-T & MW-R & MW-RW & Rc-R & Rc-Pl \\
\midrule
$\tau=0.9$ & 1/2/97 & 20/38/43 & 40/21/40 & 70/21/9 & 61/22/16 & 99/1/0 & 100/0/0 \\
$\tau=0.8$ & 8/16/76 & 59/32/8 & 77/10/12 & 89/8/2 & 84/12/4 & 100/0/0 & 100/0/0 \\
\bottomrule
\end{tabular}
\captionof{table}{\textbf{Selected newest-frame budgets.} Percentage of decisions using 128/192/256 positions. One-step-horizon decisions count as 256. Percentages are rounded independently and can sum to 99 or 101.}
\label{tab:supp_budgets}
\end{minipage}\par\vspace{8pt}

}]

\subsection{What adaptivity changes}
At $\tau=0.9$, SOAR achieves 49.26\% task-macro success versus 48.96\% for dense planning, at a $2.31\times$ speedup. Fixed 192 has the same task-macro success at a $2.21\times$ speedup. Relative to fixed 192, SOAR records seven more Push-T successes and six more Wall successes, four fewer on MW-R and five fewer on MW-RW. \Cref{tab:supp_budgets} reports the selected budgets by task and decision.

The selected budgets vary across both tasks and decisions. Maze uses 256 positions at 97\% of decisions, while MetaWorld and RoboCasa select 128 positions more frequently. Wall and Push-T use all three budgets. At $\tau=0.8$, the latency ratio is 0.394 ($2.54\times$); 77\% of Push-T decisions use 128 positions, and Push-T records eleven fewer successes than dense, compared with four fewer at $\tau=0.9$.

\subsection{Operating-point criteria and paired comparisons}
The experiment used three point-estimate criteria: a latency ratio at most 0.50, macro success at least the full-budget cached variant minus one percentage point, and Push-T success within five points of dense. At $\tau=0.9$, the respective values are 0.433, $-0.89$ points and $-4.17$ points. At $\tau=0.8$, the latter two values are $-1.93$ and $-11.46$ points, so that setting does not meet those criteria.

The paired-bootstrap differences for $\tau=0.9$, with 95\% confidence intervals, are $+0.30$ points versus dense ($[-4.17,4.91]$), $-0.89$ versus fixed 256 ($[-4.91,3.27]$), and 0.00 versus fixed 192 ($[-4.17,4.17]$).

\subsection{Development probe diagnostics}
\Cref{tab:supp_probe} reports ranking diagnostics on a separate development panel. Planning uses 256 positions, and lower-budget predictions measure rank correlation. The panel uses eight episodes on each of the first five tasks and four per RoboCasa task, all at development seed 7027.

\begin{table}[!htbp]
\centering\small\setlength{\tabcolsep}{5pt}
\begin{tabular}{lrrr}
\toprule
Task & Decisions & Median $\rho_{192}$ & Median $\rho_{128}$ \\
\midrule
Maze & 8 & 0.84 & 0.64 \\
Wall & 8 & 0.91 & 0.83 \\
Push-T & 8 & 0.87 & 0.85 \\
MW-R & 56 & 0.97 & 0.95 \\
MW-RW & 56 & 0.95 & 0.93 \\
Rc-R & 48 & 0.99 & 0.99 \\
Rc-Pl & 48 & 0.99 & 0.99 \\
\bottomrule
\end{tabular}
\caption{\textbf{Development ranking diagnostics.} Spearman correlation between reduced-budget and 256-position cached probe costs. The complete planner remains at 256 positions. These diagnostic decisions are excluded from formal success tables.}
\label{tab:supp_probe}
\end{table}

On the development panel, Push-T has a median probe-cost correlation of 0.85 at 128 positions. In the closed-loop evaluation in \cref{tab:supp_soar_detail}, fixed 128 records seventeen fewer Push-T successes than dense. The shared thresholds 0.9 and 0.8 were fixed from development probe statistics before evaluation, as described in main-paper Sec.~5.3.

\twocolumn[{
\section{Trained Sparse-Model Controls}
\label{sec:supp_trained}
\noindent\begin{minipage}{\textwidth}
\centering
\small
\setlength{\tabcolsep}{3.5pt}
\begin{tabular}{lcccccc}
\toprule
Method & Maze & Wall & Push-T & MW-R & MW-RW & Avg. \\
\midrule
{SI} & {76.50\,\scriptsize(2.12)} & {50.58\,\scriptsize(0.80)} & {47.11\,\scriptsize(4.14)} & {50.00\,\scriptsize(5.46)} & {29.28\,\scriptsize(3.30)} & {50.69} \\
{Dense-SI} & {82.75\,\scriptsize(0.53)} & {77.31\,\scriptsize(2.26)} & {66.78\,\scriptsize(1.06)} & {48.03\,\scriptsize(1.75)} & {27.78\,\scriptsize(3.67)} & {60.53} \\
{Scope} & {80.44\,\scriptsize(1.57)} & {76.74\,\scriptsize(4.86)} & {67.48\,\scriptsize(1.00)} & {47.80\,\scriptsize(4.82)} & {35.53\,\scriptsize(1.91)} & {61.60} \\
{Dense-Scope} & {82.18\,\scriptsize(1.97)} & {77.43\,\scriptsize(1.04)} & {66.78\,\scriptsize(1.40)} & {48.50\,\scriptsize(3.55)} & {28.01\,\scriptsize(3.51)} & {60.58} \\
{DDP} & {51.16\,\scriptsize(10.24)} & {76.74\,\scriptsize(3.01)} & {36.34\,\scriptsize(3.81)} & {30.56\,\scriptsize(13.68)} & {11.69\,\scriptsize(6.22)} & {41.30} \\
{Dense-DDP} & {80.79\,\scriptsize(0.72)} & {76.04\,\scriptsize(1.25)} & {67.25\,\scriptsize(0.53)} & {51.97\,\scriptsize(3.34)} & {24.77\,\scriptsize(0.72)} & {60.16} \\
\bottomrule
\end{tabular}
\captionof{table}{\textbf{Trained adaptations: success (\%).} Each task and branch has $3\times3\times96=864$ episodes. Parentheses are sample SDs across the three training-seed means, each averaged over three evaluation seeds. Each sparse method is compared only with its matched continued-training dense branch.}
\label{tab:supp_trained}
\end{minipage}\par\vspace{8pt}

}]

The trained-model comparison evaluates adaptations of Sparse Imagination (SI)~\cite{chun2026sparse}, Scope~\cite{li2026scopewm} and DDP~\cite{yin2026ddpwm} on five tasks using the CEM budget in Sec.~A.1. Each adaptation has its own continued-training dense control, giving six branches. Training seeds are 0, 42 and 2026, and evaluation seeds are 1, 101 and 1001. The 270 task--branch--training-seed--evaluation-seed combinations comprise 25,920 method episodes.

\paragraph{Training budgets and checkpoint choice.}
SI receives 10,000 updates on Maze, Wall and MetaWorld and 7,500 on Push-T. Scope includes 3,000 selector updates and 5,000 fresh-dynamics updates. DDP includes 2,000 localization, 5,000 primary-dynamics and 10,000 cumulative LRM updates. For each branch, checkpoint selection follows a fixed validation-loss criterion; closed-loop success is then measured using the selected checkpoint.

\begin{table}[H]
\centering\small\setlength{\tabcolsep}{4pt}
\begin{tabular}{lrrrr}
\toprule
Method & Success & Dense & Seconds & Speedup \\
\midrule
SI & 50.69 & 60.53 & 11.021 & 2.333$\times$ \\
Scope & 61.60 & 60.58 & 12.348 & 2.046$\times$ \\
DDP & 41.30 & 60.16 & 16.463 & 1.557$\times$ \\
\bottomrule
\end{tabular}
\caption{\textbf{Trained-control efficiency.} Success values are five-task macro percentages. Seconds averages sparse complete-planning latency over all recorded non-warm-up decisions. Speedup is the corresponding dense average divided by Seconds; the dense averages are 25.712, 25.265 and 25.625 seconds for SI, Scope and DDP, respectively.}
\label{tab:supp_train_time}
\end{table}

Scope has the highest five-task macro success among the three trained adaptations and gains 7.52 points on MW-RW against its matched dense branch. SI and DDP have lower macro success than their corresponding dense branches. The training-free experiments use frozen released checkpoints.

Dense-Scope planning latency reaches 65.531 seconds/decision on Wall at training seed 42 and 66.951 on Maze at training seed 2026. Dense-SI and Dense-DDP also have slow MW-RW runs at training seed 42.

\section{Alternative Planning and Selection Strategies}
\label{sec:supp_alternatives}
\subsection{Dense rechecking within sparse CEM}
The multi-fidelity experiment in \cref{tab:supp_multifid} keeps a 128-position checkerboard for the main search. At each CEM iteration, it recomputes selected candidate trajectories at all 256 positions, uses their full costs to select elites, and refits the CEM distribution to the elite action sequences. The fixed-query controls recheck 20 or 80 trajectories per iteration. Risk-adaptive expansion and stratified witnesses expand the full-query set in stages while retaining the same spatial mask.

\begin{table}[H]
\centering
\small
\setlength{\tabcolsep}{1.5pt}
\begin{tabular}{lcccc}
\toprule
Method & Maze & Wall & Push-T & MW-R \\
\midrule
Dense & {73.61\,\scriptsize(11.55)} & {80.90\,\scriptsize(4.20)} & {61.11\,\scriptsize(2.46)} & {54.51\,\scriptsize(4.02)} \\
Grid 50\% & {67.36\,\scriptsize(11.55)} & {76.04\,\scriptsize(2.25)} & {37.85\,\scriptsize(3.84)} & {39.93\,\scriptsize(1.30)} \\
Fixed 20 & {74.65\,\scriptsize(11.95)} & {82.64\,\scriptsize(4.68)} & {53.12\,\scriptsize(4.50)} & {39.93\,\scriptsize(4.36)} \\
Fixed 80 & {74.65\,\scriptsize(12.45)} & {78.12\,\scriptsize(3.71)} & {57.64\,\scriptsize(1.77)} & {50.69\,\scriptsize(4.36)} \\
Risk-adaptive & {75.00\,\scriptsize(12.84)} & {81.60\,\scriptsize(5.66)} & {57.29\,\scriptsize(1.70)} & {44.79\,\scriptsize(6.42)} \\
Stratified & {73.96\,\scriptsize(11.69)} & {82.64\,\scriptsize(6.03)} & {57.99\,\scriptsize(1.96)} & {49.31\,\scriptsize(5.66)} \\
\bottomrule
\end{tabular}
\par\vspace{5pt}
\begin{tabular}{lcccc}
\toprule
Method & MW-RW & Rc-R & Rc-Pl & Avg. \\
\midrule
Dense & {23.26\,\scriptsize(2.14)} & {20.83\,\scriptsize(5.89)} & {34.38\,\scriptsize(5.10)} & {49.80} \\
Grid 50\% & {17.71\,\scriptsize(4.25)} & {12.50\,\scriptsize(2.55)} & {37.50\,\scriptsize(9.20)} & {41.27} \\
Fixed 20 & {18.75\,\scriptsize(1.47)} & {12.50\,\scriptsize(4.42)} & {33.33\,\scriptsize(5.31)} & {44.99} \\
Fixed 80 & {21.88\,\scriptsize(4.50)} & {14.58\,\scriptsize(7.80)} & {34.38\,\scriptsize(2.55)} & {47.42} \\
Risk-adaptive & {20.49\,\scriptsize(3.22)} & {10.42\,\scriptsize(6.42)} & {36.46\,\scriptsize(3.90)} & {46.58} \\
Stratified & {21.18\,\scriptsize(6.03)} & {16.67\,\scriptsize(5.31)} & {31.25\,\scriptsize(4.42)} & {47.57} \\
\bottomrule
\end{tabular}
\caption{\textbf{Dense rechecking during sparse CEM: success (\%).} Seeds 2001, 3001 and 4001; 1,632 episodes per row. Parentheses give the population SD over the three evaluation seeds. All sparse searches use 128 checkerboard positions; rechecking evaluates selected trajectories at 256 positions. Fixed 20/80 denote fixed query counts; Stratified denotes stratified witnesses.}
\label{tab:supp_multifid}
\end{table}

\begin{table}[H]
\centering\small\setlength{\tabcolsep}{4pt}
\begin{tabular}{lrrr}
\toprule
Method & Success & Seconds & Ratio \\
\midrule
Dense & 49.80 & 43.538 & 1.000 \\
Grid 50\% & 41.27 & 18.629 & 0.438 \\
Fixed 20 rechecks & 44.99 & 21.833 & 0.514 \\
Fixed 80 rechecks & 47.42 & 29.466 & 0.697 \\
Risk-adaptive expansion & 46.58 & 30.065 & 0.714 \\
Stratified witnesses & 47.57 & 29.993 & 0.712 \\
\bottomrule
\end{tabular}
\caption{\textbf{Dense rechecking: latency--success trade-off.} Success is seven-task macro percent, Seconds is task-averaged complete planning time, and Ratio is the average per-task latency ratio to dense. The ratio is not obtained by dividing the Seconds column.}
\label{tab:supp_multifid_time}
\end{table}

Fixed 80 rechecks restores 6.15 macro points over Grid. Risk-adaptive expansion reaches fewer successes at a higher macro latency: 95.78\% of search rounds trigger expansion and 94.04\% reach the third stage, leaving an average of 77.11 full queries per round. Stratified witnesses reaches the third stage in 90.29\% of rounds, with only a 0.15-point mean gain over fixed 80 and slightly greater latency. Full-cost rechecking determines elite selection within CEM. The planner executes the action prefix of the refitted mean sequence without an additional full-rollout recheck.

Dense/Rc-Pl at seed 2001 takes 125.468 seconds, versus 50.686 and 50.968 for the other two seeds. The source of this timing difference has not been identified.

\subsection{Action-response token selection}
\begin{table}[H]
\centering
\small
\setlength{\tabcolsep}{2pt}
\begin{tabular}{lcccccc}
\toprule
Task & Dense & Grid & G+P & Mag. & Comp. & Shuf. \\
\midrule
Maze & 60.42 & 54.17 & 54.17 & 54.17 & 53.12 & 45.83 \\
Wall & 75.00 & 77.08 & 77.08 & 67.71 & 71.88 & 39.58 \\
Push-T & 59.38 & 42.71 & 42.71 & 51.04 & 54.17 & 15.62 \\
MW-R & 48.96 & 38.54 & 38.54 & 30.21 & 29.17 & 35.42 \\
MW-RW & 26.04 & 17.71 & 17.71 & 7.29 & 11.46 & 14.58 \\
Rc-R & 25.00 & 9.38 & 9.38 & 15.62 & 12.50 & 18.75 \\
Rc-Pl & 40.62 & 40.62 & 40.62 & 56.25 & 56.25 & 50.00 \\
Avg. & 47.92 & 40.03 & 40.03 & 40.33 & 41.22 & 31.40 \\
\bottomrule
\end{tabular}
\caption{\textbf{Action-response selection: success (\%).} Evaluation seed 3001; 544 episodes per method. G+P: Grid plus probes; Mag.: response magnitude; Comp.: response complementarity; Shuf.: shuffled correspondence. Dense and Grid are the seed-3001 controls from the rechecking batch.}
\label{tab:supp_response}
\end{table}
Nine fixed action probes describe each position's response to actions. The selectors retain 128 positions by response magnitude or complementarity; a shuffled-correspondence control breaks the association between responses and positions. This experiment contains four new methods, 28 jobs and 2,176 method episodes.

Response complementarity averages 41.22\%, versus Grid's 40.03\% and magnitude selection's 40.33\%. Its gains concentrate on Push-T (52 versus 41 successes out of 96) and Rc-Pl (18 versus 13 out of 32), with losses on both MetaWorld tasks. Its total successes are 233/544 versus Grid's 237/544, illustrating why pooled success counts and macro success can move in opposite directions.

Complementarity and magnitude selection share an average of 123.38 of their 128 positions. Shuffling the association between positions and responses reduces macro success from 41.22\% for complementarity to 31.40\%.

Complete planning latency, averaged equally across the seven tasks, is 18.716 seconds/decision for complementarity, 18.778 for magnitude selection and 18.714 for the shuffled control. Within the complementarity runs, the instrumented probe and selection times are 89.114 and 6.355 milliseconds. The mean across tasks of their combined share of complete planning time is 0.560\%. The Grid-plus-probes run reproduces the Grid actions and outcomes and averages 20.181 seconds/decision. Its reused Grid control was timed in the earlier rechecking batch and averages 18.661 seconds/decision; the MW-R timings differ by approximately $2.14\times$ between these batches.

Probe-action ranking agreement is 85.59\% for Grid, 86.77\% for magnitude, 87.15\% for complementarity and 79.34\% for shuffled correspondence. These agreements are computed on the nine probe actions used for token selection.

\section{Evaluation Coverage}
\label{sec:supp_scope}
The spatial, recovery, trained, rechecking and response experiments comprise 54,208 formal method episodes, counting reused controls once. The four temporal campaigns contribute 25,024 more, for a subtotal of 79,232. Including cached-rollout and adaptive-budget evaluations brings the study above 80,000 method episodes, as reported in the main paper. Counts refer to executed method--episode pairs; paired comparisons share initial states and goals.

The spatial, temporal and SOAR evaluations cover seven JEPA-WM tasks with the released CEM search budget. Trained sparse-model controls cover the first five tasks. Development panels provide the ranking and implementation diagnostics reported in Secs.~D and E.

\end{document}